\documentclass[10pt,letterpaper]{article}

\usepackage{cogsci}
\usepackage{pslatex}
\usepackage{apacite}

\newcommand{\remove}[1]{{\iffalse #1 \fi}}

\usepackage{amsmath}
\usepackage{graphicx}

\title{Assessing the Linguistic Productivity of Unsupervised Deep Neural Networks}
 
\author{{\large \bf Lawrence Phillips (Lawrence.Phillips@pnnl.gov)} \\
  Pacific Northwest National Laboratory
    \AND {\large \bf Nathan Hodas (Nathan.Hodas@pnnl.gov)} \\
  Pacific Northwest National Laboratory}

\begin{document}

\maketitle

\begin{abstract}

Increasingly, cognitive scientists have demonstrated interest in applying tools from deep learning. One use for deep learning is in language acquisition where it is useful to know if a linguistic phenomenon can be learned through domain-general means. To assess whether unsupervised deep learning is appropriate, we first pose a smaller question: Can unsupervised neural networks apply linguistic rules productively, using them in novel situations? 
We draw from the literature on determiner/noun productivity by training an unsupervised, autoencoder network measuring its ability to combine nouns with determiners. 
Our simple autoencoder creates combinations it has not previously encountered and produces a degree of overlap matching adults. While this preliminary work does not provide conclusive evidence for productivity, it warrants further investigation with more complex models. Further, this work helps lay the foundations for future collaboration between the deep learning and cognitive science communities.

\end{abstract}

\section{Introduction}

Computational modeling has long played a significant role within cognitive science, allowing researchers to explore the implications of cognitive theories and to discover what properties are necessary to account for particular phenomena~\cite{mcclelland2009place}. Over time, a variety of modeling traditions have seen their usage rise and fall. While the 1980s saw the rise in popularity of connectionism~\cite{thomas2008connectionist}, more recently symbolic Bayesian models have risen to prominence~\cite{chater2008probabilistic,lee2011cognitive}. While the goals of cognitive modelers have largely remained the same, increases in computational power and architectures have played a role in these shifts~\cite{mcclelland2009place}. Following this pattern, recent advances in the area of deep learning (DL) have led to a rise in interest from the cognitive science community as demonstrated by a number of recent workshops dedicated to DL~\cite{workshopCogSci2014,workshopCogSci2016,workshopNCPW15}.

As with any modeling technique, DL can be thought of as a tool which is best suited to answering particular types of questions. One such question is that of \emph{learnability}, whether an output behavior could ever be learned from the types of input given to a learner. These types of questions play an integral role in the field of language acquisition where researchers have argued over whether particular aspects of language could ever be learned by a child without the use of innate, language-specific mechanisms~\cite{smith1999infants,yang2004universal,chater2010language,pearl2014evaluating}. The success of a domain general learner does not necessarily imply that human learners acquire the phenomenon in a similar fashion, but it does open the possibility that we need not posit innate, domain-specific knowledge. 

The crux of these learning problems typically lies in making a particular generalization which goes beyond the input data. One major type of generalization that DL models would need to capture is known as \emph{linguistic productivity}. A grammatical rule is considered productive when it can be applied in novel situations. 
For example, as a speaker of English you may never have encountered the phrase \emph{a gavagai} before, but you now know that  \emph{gavagai} must be a noun and can therefore combine with other determiners to produce a phrase such as \emph{the gavagai}. Before DL might be applied to larger questions within language acquisition, the issue of productivity must first be addressed. If DL models are not capable of productivity, then they cannot possibly serve to model the cognitive process of language acquisition. On the other hand, if DL models demonstrate basic linguistic productivity, we must explore what aspects of the models allow for this productivity.

\subsection{The Special Case of Determiners}

For decades, debate has raged regarding the status of productive rules among children acquiring their native language. On the one hand, some have argued that children seem hardwired to apply rules productively and demonstrate this in their earliest speech~\cite{valian2009abstract,yang2011statistical}. On the other, researchers have argued that productivity appears to be learned, with children's early speech either lacking productivity entirely or increasing with age~\cite{pine1996syntactic,pine2013young,meylan2017emergence}. Of particular interest to this debate has been the special case of English determiners. In question is whether or not English-learning children have acquired the specific linguistic rule which allows them to create a noun phrase (NP) from a determiner (DET) and noun (N) or if they have simply memorized the combinations that they have previously encountered. This linguistic rule, $\text{NP} \rightarrow \text{DET N}$, is productive in two senses. First, it can be applied to novel nouns, e.g. \emph{a gavagai}. Second, consider the determiners \emph{a} and \emph{the}. If a singular noun can combine with one of these determiners, it may also combine with the other, e.g. \emph{the wug}.

This type of rule seems to be acquired quite early in acquisition, making it appropriate to questions of early productivity, and provides an easy benchmark for a DL model. Yet answering such a simple question first requires addressing how one might measure productivity. Most attempts to measure productivity have relied on what is known as an \emph{overlap score}, intuitively what percentage of nouns occur with both \emph{a} and \emph{the}~\cite{yang2011statistical}. This simple measure has been the source of some controversy. \citeA{yang2011statistical} argues that early attempts failed to take into account the way in which word frequencies affect the chance for a word to ``overlap". Because word frequency follows a Zipfian distribution, with a long tail of many infrequent words, many nouns are unlikely to ever appear with both determiners. He proposes a method to calculate an expected level of overlap which takes into account these facts. 
Alternatively, \citeA{meylan2017emergence} propose a Bayesian measure of productivity which they claim takes into account the fact that certain nouns tend to prefer one determiner over another. For instance, while one is more likely to hear \emph{a bath} than the phrase \emph{the bath}, the opposite is true of the noun \emph{bathroom} which shows a preference for the determiner \emph{the}~\cite{meylan2017emergence}. 

The literature is quite mixed regarding whether or not children show early productivity. Differences in pre-processing have lead researchers to draw opposite conclusions from similar data, making interpretation quite difficult~\cite{yang2011statistical,pine2013young}. Indeed, most corpora involving individual children are small enough that \citeA{meylan2017emergence} argue it is impossible to make a statistically significant claim as to child productivity. 
For analyzing whether or not text generated by a DL model is productive or not, we thankfully do not need to fully address the problem of inferring child productivity. Ideally, the model would demonstrate a similar level of overlap to the data it was exposed to. 
We make use of the overlap statistic from Yang because it is more easily comparable to other works and has been better studied than the more recent Bayesian metric of \citeA{meylan2017emergence}.

\subsection{Deep Learning for Language Acquisition}

Deep learning, or deep neural networks, are an extension of traditional artificial neural networks (ANN) used in connectionist architectures. A ``shallow" ANN is one that posits a single hidden layer of neurons between the input and output layers. Deep networks incorporate multiple hidden layers allowing these networks in practice to learn more complex functions. The model parameters can be trained through the use of the backpropogation algorithm. The addition of multiple hidden layers opens up quite a number of possible architectures, not all of which are necessarily applicable to problems in cognitive science or language acquisition more specifically.

While the most common neural networks are discriminative, i.e. categorizing data into specific classes, a variety of techniques have been proposed to allow for truly generative neural networks. These generative networks are able to take in input data and generate complex outputs such as images or text which makes them ideal for modeling human behavior. We focus on one generative architecture in particular known as a deep \emph{autoencoder} (AE)~\cite{hinton2006reducing}. 

While AEs have been used for a variety of input data types, most prominently images, we describe their use here primarily for text. The first half, the \emph{encoder}, takes in sentences and transforms them into a condensed representation. This condensed representation is small enough that the neural network cannot simply memorize each sentence and instead is forced to encode only the aspects of the sentence it believes to be most important. The second half, the \emph{decoder}, learns to take this condensed representation and transform it back into the original sentence. Backpropogation is used to train model weights to reduce the loss between the original input and the reconstructed output. 
Although backpropagation is more typically applied to supervised learning problems, the process is in fact unsupervised because the model is only given input examples and is given no external feedback. 

AEs have been shown to successfully capture text representations in areas such as paragraph generation~\cite{li2015hierarchical}, part-of-speech induction~\cite{vishnubhotla2010autoencoder}, bilingual word representations~\cite{ap2014autoencoder}, and sentiment analysis~\cite{socher2011semi}, but have not been applied to modeling language acquisition. While any number of DL architectures could be used to model language acquisition, the differences between ANNs and actual neurons in the brain make any algorithmic claims difficult. Instead, DL models might be used to address computational-level questions, for instance regarding whether or not a piece of knowledge is learnable from the data encountered by children. Before this can be done, however, it remains to be seen whether DL models are even capable of creating productive representations. If they cannot, then they do not represent useful models of language acquisition.
This work attempts to address this not by creating a model of how children acquire language, but by using methods from the psychological literature on productivity to assess the capability of DL to learn productive rules.

\section{Methods}

\subsection{Corpora}
To train our neural network, we make use of child-directed speech taken from multiple American-English corpora in the CHILDES database~\cite{macwhinney2000childes}. In particular, we make use of the CDS utterances in the Bloom 1970, Brent, Brown, Kuczaj, Providence, Sachs, and Suppes corpora~\cite{bloom70,Brent,Brown,Kuczaj,Providence,Sachs,Suppes}. The combined corpora contain almost 1 million utterances and span a wide age range, including speech directed to children as young as 6 months and as old as 5 years. Relevant information about the used corpora can be found in Table~\ref{tab:corpora}.

Because we are interested in seeing what the AE can learn from data similar to that encountered by children, we train the model only on child-\emph{directed} utterances. These can be produced by any adult in the dataset, including parents and researchers. Although a comparison with child-produced text holds great interest, it is not clear whether child-produced speech is rich enough to support robust language learning on its own. It therefore provides a poor basis upon which to train the AE.

Text from the various corpora is processed as a single document. Child-directed utterances are cleaned from the raw files using the CHILDESCorpusReader function of the Python Natural Language Toolkit (NLTK). Utterances from all non-children speakers are included and not limited just to the primary caregiver. Each utterance is split into words according to the available CHILDES transcription and then made lowercase. The model represents only the most frequent 3000 words, while the remainder are represented as a single \emph{out-of-vocabulary} (OOV) token. This step is taken both to reduce computational complexity but also to mimic the fact that young children are unlikely to store detailed representations of all vocabulary items encountered. Because the neural networks require each input to be of the same length, sentences are padded to a maximum length of 10 words. Sentences that are longer than this are truncated, while short sentences are prepended with a special \emph{PAD} token.

\begin{table}[h]
\centering
\begin{tabular}{l r r}
{\bf Corpora} & {\bf Age Range} & {\bf N. Utterances} \\ \hline
Bloom 1970 & 1;9 - 3;2 & 62,756 \\
Brent & 0;6 - 1;0 & 142,639 \\
Brown & 1;6 - 5;1 & 176,856 \\
Kuczaj &2;4 - 4;1 & 57,719\\
Providence &1;0 - 3;0 & 394,800 \\
Sachs & 1;1 - 5;1& 28,200 \\
Suppes &1;11 - 3;3 & 67,614 \\ \hline
Overall & 0;6 - 5;1 & 930,584 \\ \hline
\end{tabular}
\caption{\label{tab:corpora} Descriptive statistics of CHILDES corpora. Ages are given in (year;month) format and indicate the age of the child during corpus collection.}
\end{table}

\subsection{Neural Network Architecture}

\begin{figure}
\centering
\includegraphics[width=1\linewidth]{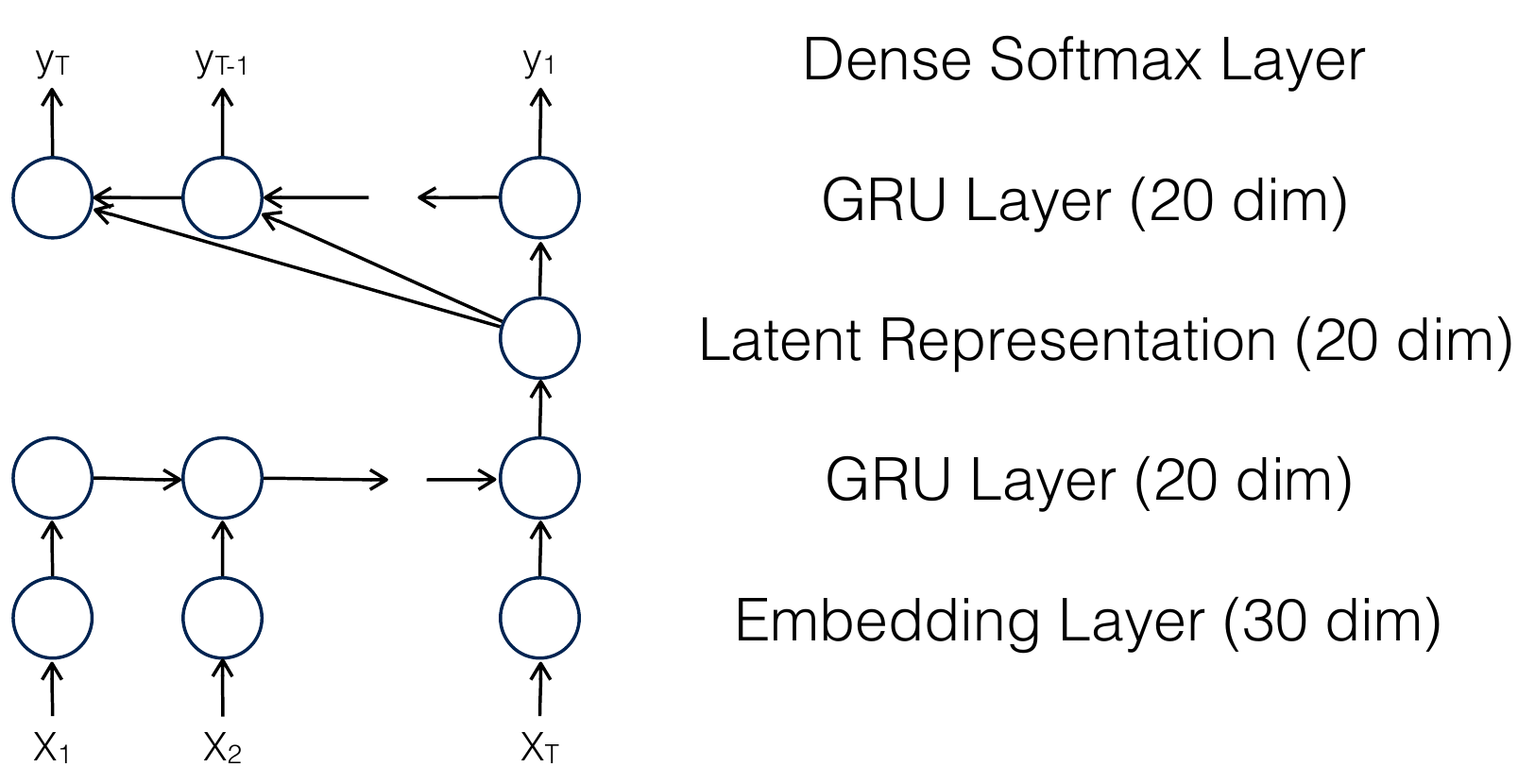}
\caption{\label{fig:autoencoder} Visual representation of the autoencoder model.}
\end{figure}

Our autoencoder model was implemented using Keras and Tensorflow. The words in each sentence are input to the model as a one-hot vector, a vector of 0s with a single 1 whose placement indicates the presence of a particular word. This is an inefficient representation because it assumes all words are equally similar, e.g. that \emph{dog} is equally similar to \emph{dogs} as it is to \emph{truck}. To deal with this, the model passes the one-hot vector to an embedding layer. Neural word embeddings, as popularized by the word2vec algorithm~\cite{mikolov2013efficient}, are a way to represent words in a low-dimensional space without requiring outside supervision. Words are placed within the space such that words that are predictive of neighboring words are placed closer to one another. 
Because our training data is relatively small, we keep the embedding dimensionality low, at only 30. Standard embeddings trained on much larger NLP corpora tend to use 100 or 200 dimensions.

Once each word has been transformed into a 30-dimensional embedding  vector, the sequence of words is passed into a gated-recurrent unit (GRU) layer~\cite{cho2014learning}. The GRU is a type of recurrent (RNN) layer which we choose because it can be more easily trained. RNN layers read in their inputs sequentially and make use of hidden ``memory" units that pass information about previous inputs to later inputs, making them ideal for sequence tasks such as language. As such, the model creates a representation of the sentence which it passes from word to word. The final representation is the output of the encoder, a latent representation of the full sentence.

This 20-dimensional latent vector serves as the input to the decoder unit. The first layer of the decoder is a GRU layer of the same shape as in the encoder. For each timestep, we feed into the GRU the latent vector, similar to the model proposed in \citeA{cho2014learning}. Rather than producing a single output, as in the encoder, the decoder's GRU layer outputs a vector at each timestep. Each of these vectors is fed into a shared dense softmax layer which produces a probability distribution over vocabulary items. The model then outputs the most likely word for each timestep.

The model loss is calculated based on the model's ability to reconstruct the original sentence through categorical crossentropy. Model weights are trained using the Adam optimzer over 10 epochs. During each epoch the model sees the entire training corpus, updating its weights after seeing a batch of 64 utterances. While this process does not reflect that used by a child learner, it is a necessary component of training the neural network on such a small amount of data. If the network had access to the full set of speech that a child encounters such a measure likely would not be necessary. Future work might also investigate whether optimizing the dimensionality of the network might lead to better text generation with higher levels of productivity.

\subsection{Baseline Models}
Because the AE is learning to reproduce its input data, one might wonder whether similar results might be achieved by a simpler, distributional model. To assess this, we also measure the performance of an n-gram language model. We train bigram and trigram language models using the modified Kneser-Ney smoothing \cite{heafield2013scalable}  implemented in the KenLM model toolkit to estimate the distributional statistics of the training corpus. Sentences are generated from the n-gram language model by picking a seed word and then sampling a new word from the set of possible n-grams. The smoothing process allows for the model to generate previously unseen n-grams. Sampling of new words continues for each utterance until the end-of-sentence token is generated or a maximum of 10 tokens is reached (the same maximum size as for the AE).

Since the AE is able to generate sentences from a latent representation, it would be inappropriate to generate n-gram sentences from random seed words. Instead, for every sentence in the test set we begin the n-gram model with the first word of the utterance. While this allows the model to always generate its first token correctly, this does not directly impact our measure of productivity as it relies on combinations of tokens. 

\subsection{Productivity Measures}

We measure the productivity of our autoencoders through the overlap score described in \citeA{yang2011statistical}. Words both in the child-directed corpus and the autoencoder-generated output are tagged using the default part-of-speech tagger from NLTK. The empirical overlap scores are simply calculated as a percentage of unique nouns that appear immediately after both the determiners \emph{a} and \emph{the}. The expected overlap score is calculated based off of three numbers from the corpus under consideration, the number of unique nouns $N$, the number of unique determiners $D$, and the total number of noun/determiner pairs $S$. The expected overlap is defined as in Equation \ref{eq:expOverlap}:

\begin{equation}
\label{eq:expOverlap}
O(N,D,S)  = \frac{1}{N} \sum_{r=1}^{N} O(r,N,D,S)
\end{equation}

where $O(r,N,D,S)$ is the expected overlap of the noun at frequency rank $r$:

\begin{equation}
\label{eq:expOverlap2}
O(r, N, D, S) = 1 + (D-1)(1-p_r)^S - \sum_{i=1}^{D}[(d_ip_r+1-p_r)^S]
\end{equation}

$d_i$ represents the probability of encountering determiner $i$, for which we use the relative frequencies of \emph{a} and \emph{the} calculated from the training corpus (39.3\% and 60.7\%, respectively). The probability $p_r$ represents the probability assigned to a particular word rank. The Zipfian distribution takes a shape parameter, $a$ which \citeA{yang2011statistical} set equal to 1 and which we optimize over the training corpus using least squares estimation and set at 1.06:

\begin{equation}
\label{eq:expOverlap3}
p_r = \frac{^{1}/_{r^a}}{\sum_{n=1}^N (\frac{1}{n^a})}
\end{equation}

It should be noted that Zipfian distributions are not perfect models of word frequencies~\cite{piantadosi2014zipf}, but assigning empirically-motivated values to the determiner probabilities and Zipfian parameter $a$ represents an improvement upon the original measure.

\section{Results}

\begin{figure}
\centering
\includegraphics[width=\linewidth]{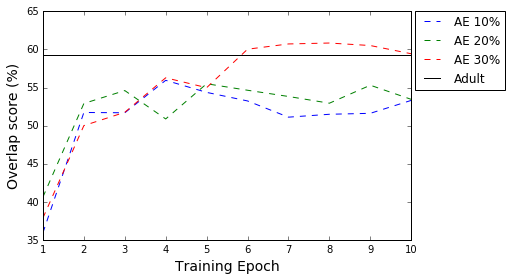}
\caption{\label{fig:overlap} Empirical overlap scores. Adult-generated speech is marked by the solid black line while autoencoder-generated speech is marked by the dashed colored lines. Results are presented for three levels of dropout, 10\%, 20\%, and 30\%. The x-axis represents the training epoch of the model.}
\end{figure}

We analyze our overlap measures for the adult-generated (i.e. child-directed) as well as the autoencoder and n-gram model-generated text and present these results in Figure \ref{fig:overlap}. We analyze overlap scores across 10 training epochs with three levels of dropout, 10\%, 20\%, and 30\%. Dropout is typically included in neural models to encourage the model to better generalize. We hypothesized that a certain level of dropout would encourage the model to generate novel combinations of words that might lead to higher overlap scores. We find that with only two training epochs the AEs have already begun to near their maximum overlap performance. The 30\% dropout AE achieves the highest level of performance, matching the empirical overlap score of the original corpus. The 10\% and 20\% dropout models perform somewhat worse suggesting that high levels of dropout may be necessary for good text generation.

In Table \ref{tab:overlap}, we present the results for the final epoch of the AE models as well as for the adult-generated and n-gram generated text. We note that the expected overlap measure consistently overestimates the productivity of all learners, including the adult-generated text. It is unclear why this should be the case, but could be a result of capping the model vocabularies, resulting in lower $N$ values. In particular, the autoencoders tend to produce a relatively limited set of nouns. Looking at empirical overlap measures, the worst-performing models are the bigram and trigram models with overlap scores below 30\%. The AEs fair much better all producing overlap scores over 50\%. The 30\% dropout AE is actually able to match the overlap score of the original adult-generated corpus (59.4\% vs. 59.3\%). 

Looking at the number of unique nouns following a determiner ($N$) and the total number of determiner-noun pairs ($S$), it becomes clear there are large differences between the n-gram and AE models. The n-gram models tend to produce very few determiner-noun pairs (low $S$) but are likely to choose from any of the nouns in the corpus, leading to high $N$. This fact accounts for the low overlap scores that they achieve. In contrast, the AEs follow a pattern which mirrors the adult corpus with few unique nouns but a large number of noun-determiner pairs. In all cases, however, the AEs produce both fewer unique nouns and fewer noun-determiner pairs than the original corpus.

\begin{table}[h]
\centering
\begin{tabular}{l r r r r}
&  {\bf {\em N}}&{\bf {\em S}} & {\bf Exp. Over.} & {\bf Emp. Over.} \\ \hline
{\bf Adult} & 1,390 & 34,138 & 77.5\% & 59.3\% \\ \hline
{\bf AE 10\%} & 861 & 29,497 & 88.4\% & 53.3\% \\ 
{\bf AE 20\% }& 870 & 28,817 &  87.6\% & 53.4\% \\
{\bf AE 30\% }& 816 & 31,181 & 90.8\% & 59.4\% \\ \hline
{\bf Bigram }& 1,780 & 5,177 & 17.6\% & 28.6\% \\
{\bf Trigram } & 2,506 & 4,595 & 11.2\% & 22.1\% \\ \hline
\end{tabular}
\caption{\label{tab:overlap} Expected and empirical overlap scores for adult- and autoencoder-generated language with varying levels of dropout. Expected overlap scores were calculated as in Yang (2011). Empirical overlap was calculated as the percent of unique nouns that appeared immediately following both \emph{a} and \emph{the}.}
\end{table}

One possible problem for calculating the expected overlaps comes from the difficulty of part-of-speech tagging text generated by the neural network. Whereas adult-generated speech follows set patterns that machine taggers are built to recognize, the neural network does not necessarily generate well-formed language. Examples of AE-generated text can be found in Table \ref{tab:exampleOutput}. In some cases, the tagger treats items that occur after a determiner as a noun regardless of its typical usage. For example, in the generated sentence \emph{let put put the over over here}, the phrase \emph{the over} is tagged as a DET$+$N pair. These type of errors are further evidenced by the fact that the trigram language model produces a larger set of words tagged as nouns than the original adult-generated corpus (2,506 vs. 1,390). 

Another explanation for the difference between expected and empirical overlaps may come from deviation from a true Zipfian distribution of word frequencies. If word frequencies are Zipfian, we should expect a perfect correlation between log ranks and log counts. \citeA{yang2011statistical} report a correlation of 0.97, while our larger corpus deviates from this with $r^2 = 0.86$. Although we attempt to take this into account by fitting the Zipfian distribution's shape parameter, this divergence clearly indicates that further work is needed.

The success of the AE model in generating productive text serves as a confirmation that unsupervised neural models might be used in future work to investigate other cognitive phenomena. This work does not directly address the question of how infants might learn to produce productive speech, it does represent one possible approach. AEs can, for instance, be thought of as information compression algorithms which learn to represent high-dimensional data into a low-dimensional latent space~\cite{hinton2006reducing}. If the brain likewise attempts to find efficient representations of the stimuli it encounters then it may prove fruitful to investigate how these representations compare to one another.

\begin{table}[h]
\centering
\begin{tabular}{l l}
{\bf Adult} & {\bf Autoencoder} \\ \hline
falling down & down down \\ \hline
you're playing with & you're playing with \\
your bus &the head \\ \hline
why did OOV say what's & what what you say say  \\
wrong with these apples & say with {\bf the dada} \\
\end{tabular}
\caption{\label{tab:exampleOutput} Example adult and AE-generated language. The AE-generated text is from the final epoch of the AE with 20\% dropout. In bold is a DET$+$N combination that does not appear in the AEs input.}
\end{table}

\section{Conclusion}

While there is great interest regarding the inclusion of deep learning methods into cognitive modeling, a number of major hurdles remain. For the area of language acquisition, deep learning is poised to help answer questions regarding the learnability of complex linguistic phenomena without access to innate, linguistic knowledge. Yet it remains unclear whether unsupervised versions of deep learning models are capable of capturing even simple linguistic phenomena. In this preliminary study, we find that a simple autoencoder with sufficient levels of dropout is able to mirror the productivity of its training data, although it is unclear whether this proves productivity in and of itself.

Future work will need to investigate whether more complex models might be able to generate text with higher productivity as well as further investigating how particular model choices impact performance. It would also be worthwhile to compare AEs against simpler models such as a basic LSTM language model. While additional work needs to be done to motivate the use of deep learning models as representations of how children might learn, this preliminary work shows how one might combine techniques from deep learning and developmental psychology.

\section{Acknowledgments}

The authors thank the reviewers for their thoughtful comments and Lisa Pearl for initial discussion regarding productivity.

\bibliographystyle{apacite}

\setlength{\bibleftmargin}{.125in}
\setlength{\bibindent}{-\bibleftmargin}

\bibliography{deeplearning}

\end{document}